\documentclass{bmvc2k}

\usepackage{times}
\usepackage{epsfig}
\usepackage{graphicx}
\usepackage{amsmath}
\usepackage{amssymb}
\usepackage{graphicx,subfigure}
\usepackage{pifont}
\usepackage{xr}

\usepackage{multirow}
\usepackage{booktabs}
\usepackage[mathscr]{eucal}

\usepackage{algorithm}
\usepackage{algpseudocode}

\usepackage{tikz}
\usepackage{pgfplots}
\usetikzlibrary{positioning,arrows,arrows.meta,fit,backgrounds,calc}

\let\originalleft\left
\let\originalright\right
\renewcommand{\left}{\mathopen{}\mathclose\bgroup\originalleft}
\renewcommand{\right}{\aftergroup\egroup\originalright}

\usepackage{amsmath}
\usepackage{amssymb}  

\usepackage{mathtools}  

\makeatletter
\newcommand{\spx}[1]{
	\if\relax\detokenize{#1}\relax
	\expandafter\@gobble
	\else
	\expandafter\@firstofone
	\fi
	{^{#1}}%
}
\makeatother

\usepackage{stmaryrd}  
\newcommand{\genericdel}[4]{%
	\ifcase#3\relax
	\ifx#1.\else#1\fi#4\ifx#2.\else#2\fi\or
	\bigl#1#4\bigr#2\or
	\Bigl#1#4\Bigr#2\or
	\biggl#1#4\biggr#2\or
	\Biggl#1#4\Biggr#2\else
	\left#1#4\right#2\fi
}

\newcommand{\cbr}[2][-1]{\genericdel\{\}{#1}{#2}}

\newcommand{\intoo}[2][-1]{\mathinner{\genericdel(){#1}{#2}}}
\newcommand{\intcc}[2][-1]{\mathinner{\genericdel[]{#1}{#2}}}
\newcommand{\intoc}[2][-1]{\mathinner{\genericdel(]{#1}{#2}}}

\newcommand{\envert}[2][-1]{\genericdel\vert\vert{#1}{#2}}

\usepackage{bm}
\usepackage[OMLmathsfit]{isomath}  
\usepackage{upgreek}

\DeclareMathAlphabet{\mathbbmrm}{U}{bbm}{m}{rm}
\DeclareMathAlphabet{\mathbbmsl}{U}{bbm}{m}{sl}
\DeclareMathAlphabet{\mathbbmb}{U}{bbm}{b}{it}
\DeclareMathAlphabet{\mathbbmssit}{U}{bbmss}{m}{it}

\let\vec\relax

\newcommand{\vec}[1]{\bm{#1}}

\newcommand{\cvec}[1]{\mathbf{#1}}

\newcommand{\nunder}[2][5]{\mathrlap{\mkern\the\numexpr#1/2mu\relax\underline{\phantom{\mathrm{#2}\mkern-#1mu}}}\mathrm{#2}}
\newcommand{\subline}[2][n]{%
    \makebox[\widthof{$#2$}/2]{}\mathclap{#2}%
    \text{\smash{\raisebox{0.04em}{
        $\mathclap{\underline{\hphantom{#1}\vphantom{#2}}}$%
    }}}%
    \makebox[\widthof{$#2$}/2]{}%
}

\newcommand{\rvarstyle}[1]{{{\subline[i]{#1}}}}

\newcommand{\rvec}[1]{\rvarstyle{\vec{#1}}}

\DeclareMathOperator*{\argmax}{arg\,max}
\DeclareMathOperator*{\E}{I\kern-.282em E}

\newcommand{\enbbracket}[1]{{\mathinner{\left\llbracket{#1}\right\rrbracket}}}

\let\oldhat\hat
\renewcommand{\hat}[1]{\vphantom{#1}\smash[t]{\oldhat{#1}}}
\let\oldtilde\tilde
\renewcommand{\tilde}[1]{\vphantom{#1}\smash[t]{\oldtilde{#1}}}
\let\oldwidetilde\widetilde
\renewcommand{\widetilde}[1]{\vphantom{#1}\smash[t]{\oldwidetilde{#1}}}

\newcommand{\bidot}{\mkern1.5mu{..}\mkern1.5mu}

\newenvironment{talign*}
{\csname align*\endcsname}
{\endalign}

\usepackage{dashbox}%

\newcommand{\delete}[1]{}
\newcommand{\todelete}[1]{{\color[rgb]{0.8,0.2,0.0}{#1}}}
\newcommand{\cmark}{\ding{51}}%
\newcommand{\xmark}{\ding{55}}%

\renewcommand{\paragraph}[1]{\smallskip\noindent\textbf{#1}}

\title{Backdoor Defense through Self-Supervised and Generative Learning}

\addauthor{Ivan Sabolić}{ivan.sabolic@fer.hr}{1}
\addauthor{Ivan Grubišić}{ivan.grubisic@fer.hr}{1}
\addauthor{Siniša Šegvić}{sinisa.segvic@fer.hr}{1}

\addinstitution{
Faculty of Electrical Engineering \\
and Computing \\
University of Zagreb
}

\runninghead{Sabolić, Grubišić, Šegvić}{\small{Generative self-supervised backdoor defense}}

\begin{document}

\maketitle
\begin{abstract}
Backdoor attacks change a small portion of training data by introducing hand-crafted triggers and rewiring the corresponding labels towards a desired target class. Training on such data injects a backdoor which causes malicious inference in selected test samples. Most defenses mitigate such attacks through various modifications of the discriminative learning procedure. In contrast, this paper explores an approach based on generative modelling of per-class distributions in a self-supervised representation space. Interestingly, these representations get either preserved or heavily disturbed under recent backdoor attacks. In both cases, we find that per-class generative models allow to detect poisoned data and cleanse the dataset. Experiments show that training on cleansed dataset greatly reduces the attack success rate and retains the accuracy on benign inputs.

\end{abstract}

\section{Introduction}
Deep models are establishing themselves as 
the default approach for resolving diverse problems across various domains~\cite{he2016deep, xiong2016achieving, silver2016mastering}. 
However, their large 
capacity leaves them vulnerable to various cybernetic attacks~\citep{biggio18pr}. 
Backdoor attacks 
inject vulnerabilities into production models 
by introducing subtle changes
to the training data~\citep{gu2019badnets}.
The installed backdoor 
induces malicious behaviour
according to the attacker's goals~\cite{laskov10ml}.
We focus on backdoors that manipulate the model predictions
in presence of a pre-defined trigger~\cite{chen2017targeted}.
These attacks typically 
seek stealthiness
through unobtrusive trigger designs, 
low prevalence of poisoned data,
and high generalization performance 
on benign samples~\cite{gu2019badnets}.

Backdoor attacks must trade-off 
stealthiness with applicability.
Localized attacks can be easily 
applied in the physical world,
however they can be uncovered
by careful visual inspection~\cite{gu2019badnets, chen2017targeted}
and defenses that specialize for such attacks~\citep{salman2022certified,wang2019neural}.
Pervasive attacks may be 
imperceptible to the human eye
but they can not be applied with a sticker~\citep{nguyen2021wanet, li2021invisible}. 
Many early defenses 
are designed to counter specific types of 
attacks~\citep{wang2019neural, qiao2019defending}.
More recent empirical defenses focus on detecting 
a wide range of attacks by targeting common 
weaknesses~\citep{huang2022backdoor, chen2022effective, gao2023backdoor}.



This work proposes a novel approach  
to prevent the backdoor deployment
given potentially poisoned data 
from an untrusted source~\cite{wang2019neural}. 
We conjecture that a backdoor defense
has a better chance of success
if it 
relies more on class-agnostic features rather than
discriminative features derived from potentially poisoned labels.
However, very few previous works
consider self-supervised~\cite{huang2022backdoor, wang2023training}
or generative approaches~\cite{qiao2019defending},
while their synergy is completely unexplored.
We attempt to fill that gap 
by formulating our defense 
in terms of per-class densities
of self-supervised representations.
Our objective is to identify poisoned samples,
restore the correct labels and to produce a clean model.

Poisoned samples are generated by either 
i) injecting triggers into images of non-target classes~\citep{gu2019badnets, chen2017targeted, qi2022revisiting, li2021invisible, jiang2023color} 
or 
ii) applying strong perturbations
to images of the target class~\citep{turner2019label, barni2019new}. 
In both cases, our analysis shows that the self-supervised embeddings of poisoned samples 
get placed outside of the target class manifold. 
Hence, one could hypothesize that 
a generative model of the target class
should assign these samples a lower likelihood.
However, generative modelling of RGB images 
may assign high densities to outlier images~\citep{nalisnick2018deep, serra2019input}.
To avoid that, 
we decide to model per-class densities
in the latent space 
of a self-supervised feature extractor. 
We identify potentially poisoned data 
according to the following two tests.
The first test identifies samples 
residing within the distribution 
of a class that differs from their label. 
The second test identifies samples 
that are outside of the distribution 
of the entire dataset. 
These two tests allow us 
to cleanse the dataset
by removing suspicious samples. 
Subsequently, we restore the original labels of those samples 
through generative classification \cite{mackowiak2021generative}.
Training on cleansed data ensures high performance
on benign inputs, while fine-tuning on relabeled 
data ensures the complete removal of the backdoor.

To summarize, our contributions encompass 
three key aspects. 
First, we identify
effects of various backdoor attacks
on self-supervised image representations. 
Second, we propose a novel backdoor defense 
that builds upon per-class 
distributions of self-supervised representations. 
Third, we improve our defense 
through fine-tuning on pseudo-labels
obtained by our generative classifier.
Our experiments demonstrate  
the effectiveness of our approach 
in comparison to several 
state-of-the-art defenses.
Importantly, our method successfully 
defends against a variety of attack types, 
including 
the latest attacks designed
to undermine
defenses based on latent separability~\citep{qi2022revisiting}.
Our code is publicly available at \url{https://github.com/ivansabolic/GSSD}.

\section{Related Work}

The main goal of existing backdoor attacks 
is to increase the attack success rate~\citep{li2022backdoor}, 
while retaining stealthy triggers,
low poisoning rates and clean accuracy~\citep{gu2019badnets}. 
A variety of triggers has been introduced, 
including black-white checkerboards~\citep{gu2019badnets}, 
blending backgrounds~\citep{chen2017targeted},
invisible noise~\cite{li2020invisible}, 
adversarial patterns~\citep{zhao2020clean} 
and sample-specific patterns~\citep{li2021invisible, nguyen2020input}.
Existing attacks can further be divided into 
poisoned-label and clean-label types. 
Poisoned-label approaches~\citep{gu2019badnets, chen2017targeted, nguyen2021wanet, li2021invisible, nguyen2020input, wu2023computation, jiang2023color}
connect the trigger with the target class 
by relabeling poisoned samples as target labels. 
Clean-label approaches modify 
samples from the target class 
while leaving the labels unchanged~\cite{turner2019label, barni2019new}. 
However, they are less effective than poisoned-label attacks~\citep{li2022backdoor}.
Many existing defenses can be classified
into three categories: 
i) detection-based defenses 
~\citep{tran2018spectral, chen2018detecting, xu2021detecting, guo2023scale, khaddaj2023rethinking, huang2023distilling},
ii) training-time defenses 
~\citep{li2021neural, li2021anti, huang2022backdoor, gao2023backdoor, zhang2023backdoor, liu2023beating, jin2021incompatibility}
and iii) post-processing defenses~\citep{liu2018fine, wang2019neural, dong2021black, tao2022better, zhao2020bridging, zhu2024neural, xiang2024cbd, xu2023towards, wei2024shared, min2024towards, li2023reconstructive, pang2023backdoor, mu2023progressive}.
The goal of detection-based defenses 
is to discover poisoned samples 
in order to deny their impact. 
Training-time defenses aim 
to develop a clean model 
from a potentially poisoned dataset.
Post-processing defenses
intend to remove the backdoor
from an already trained model.

A significant drawback of detection-based defenses 
is the unused potential 
of the detected suspicious samples.
On the other hand, training-time defenses
remain vulnerable to the retained poisoned samples.
We address these limitations
by detecting poisoned data and correcting their labels
through robust inference.
After this intervention, the triggers act as data augmentation rather than an instrument for 
backdoor deployment. 
This effectively prevents the model from learning 
the association between the trigger and the target class.

\section{Motivation}

Backdoor attacks 
pose a great challenge
since the attackers hold 
the first-mover advantage~\citep{gu2019badnets}.
Nevertheless, we know that 
triggers must not disturb 
image semantics
in order to promote stealthiness~\citep{li2022backdoor}.
We propose to take advantage of this constraint
by grounding our defense on image content,
while avoiding poisoned labels.

\paragraph{Embedding inputs into the latent space.}
%
%
%
We avoid standard supervised learning
due to its tendency to learn shortcut associations~\cite{geirhos2020shortcut}
between triggers and target labels.
We start by self-supervised learning 
of a class-agnostic representation
of the training data~\cite{chen2020simple}.
Self-supervised representations exhibit remarkable semantic power, often surpassing supervised representations in linear evaluation~\cite{he2022masked, oquab2023dinov2}. 
Furthermore, they are resilient 
to poisoned-label attacks
due to strong augmentations and contrastive
learning objectives
that disregard the labels~\cite{huang2022backdoor, wang2023training}.

\begin{figure*}[htb!]
  \centering
  \includegraphics[width=0.49\textwidth] {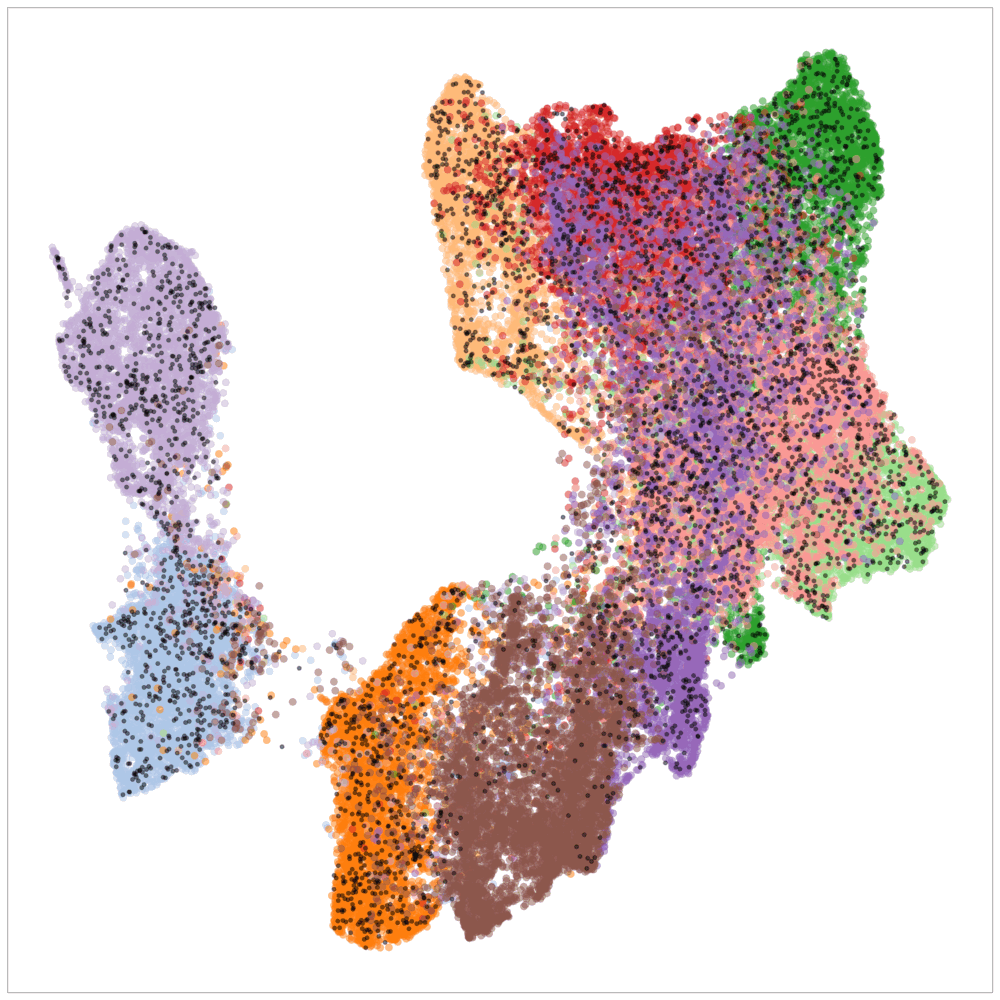}
  \includegraphics[width=0.49\textwidth]{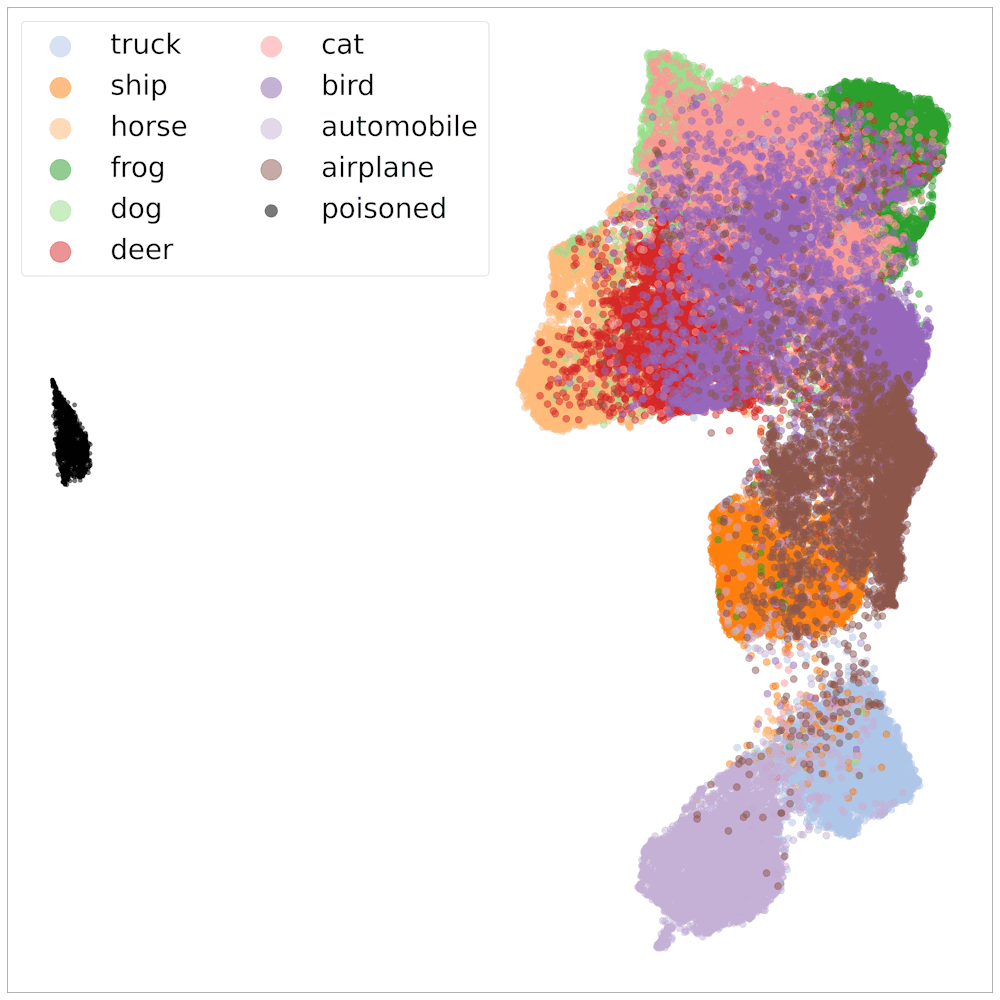}
  \caption{2D UMAP visualization 
    of the self-supervised feature space for CIFAR-10.
    Poisoned samples are shown in black, 
    while clean samples are shown in colour. 
    The target class is in brown (\textit{airplane}).
    Non-disruptive attacks 
    (left,~\cite{gu2019badnets})
    exert a very small influence 
    to the self-supervised embeddings.
    Disruptive attacks 
    (right,~\cite{turner2019label})
    displace the poisoned samples from the manifold of the training data.
  }
  \label{fig:feature_space}
\end{figure*}

\paragraph{Attack impact in the self-supervised feature space.}
\label{sub:behaviour_feature_space}
Our analysis builds upon 
UMAP~\cite{mcinnes2018umap} dimensionality reduction
that is optimized to retain 
the adjacency structure
of the original high-dimensional space. 
The resulting two-dimensional plots
show the clean data-points 
in the colours of their respective classes
and the poisoned data in black,
as shown in Figure~\ref{fig:feature_space}.

Figure~\ref{fig:feature_space} (left) illustrates 
the common behaviour 
of a large groups of attacks
from the literature~~\citep{gu2019badnets, chen2017targeted, nguyen2021wanet, qi2022revisiting}.
The plot suggests that these attacks
exert a very small influence onto 
the embedding of the clean images. 
Appendix~\ref{app:attack_impact} supports this hypothesis
with quantitative measurements.
We refer to this scenario 
as non-disruptive poisoning:
poisoned embeddings resemble their original class 
more than the target class. 
This behaviour is not unexpected 
since the triggers are designed 
for minimal visual impact 
in order to conceal the attack.


Figure~\ref{fig:feature_space} (right) shows that some attacks 
move the poisoned embeddings
off the natural manifold of the clean embeddings 
in the self-supervised feature space. 
This separation can occur 
due to adversarial triggers \cite{goodfellow2014generative} 
or other strong perturbations 
of the original images \cite{szegedy2013intriguing, turner2019label}. 
Consequently, we refer to this scenario 
as disruptive poisoning. 
One such example is the clean-label attack, 
which aims to induce an association 
between the trigger and 
the unchanged original label 
by adversarially perturbing the image 
before adding the trigger~\cite{turner2019label}.

\paragraph{Modeling per-class distributions.}
In both scenarios 
from the previous subsection,
the poisoned embeddings are situated
far away from the clean samples of the target class.
We propose to expose this occurrence
by comparing per-class densities
of the latent embeddings.
These densities can be recovered
by learning per-class generative models
such as normalizing flows
\citep{rezende15icml}
or variational encoders
\citep{kingma14iclr}.
Moreover, in the non-disruptive scenario, 
one can try to recover the original label
of poisoned samples 
through generative classification.

\section{Defense through generative and self-supervised learning}

Our defense identifies target classes
and suspicious samples by leveraging 
per-class densities 
of self-supervised features.
We re-train robust supervised models 
without suspicious samples
and subsequently fine-tune  
with corrected labels.
For brevity, we refer to our defense as GSSD (generative self-supervised defense).
We explain the details
within this section.  

\subsection{Problem formulation}
\label{subsec:formulation}
\paragraph{Threat model.}
We assume the standard \textit{all-to-one} 
threat model~\citep{gao2023backdoor, huang2022backdoor}.  
The attack poisons a subset of the original benign training dataset
$\mathcal D = \left\{(\vec{x}_i^*, y_i^*)\right\}_{i=1}^N \subset \mathcal X\times\mathcal Y$,
where $\mathcal X$ contains all possible inputs and $\mathcal Y$ is the
set of classes.
The attack involves a single target label $y_\text{T} \in \mathcal Y$
and modifies samples with indices $\mathcal I_\text{P} \subset \{1\bidot N\}$ with some transformation $(\vec x_i^*, y_i^*) \mapsto (\tilde{\vec x}_i, y_\text{T})$ to produce the poisoned subset
$\mathcal D_\text{P} =
    \cbr{
        (\tilde{\vec x}_i,
        y_\text{T})}_{i\in\mathcal I_\text{P}}$.
The complete poisoned dataset is $\tilde{\mathcal D} = \mathcal D_\text{C} \cup \mathcal D_\text{P}$, where the clean subset $\mathcal D_\text{C}$ contains the remaining (non-poisoned) samples.
The attack aims for a model trained on $\tilde{\mathcal D}$ to classify triggered test inputs as the target class $y_\text{T}$ without affecting the performance on clean inputs.


\paragraph{Defender's goals.}
We assume that the defender controls the training process. 
Given a possibly poisoned training set $\tilde{\mathcal D}$, the defender's
objective is to obtain a trained model instance without a backdoor while preserving high accuracy on benign samples.
Our defense aims to achieve this goal by producing a filtered clean subset $\hat{\mathcal D}_\text{C}$ and a relabeled poisoned subset $\hat{\mathcal D}_\text{P}'$ that can be used to train the classifier safely.

\subsection{Defense overview}
The input to our method is 
a potentially poisoned dataset: 
$\tilde{\mathcal D} \subset \mathcal X \times \mathcal Y$.
As outlined in Algorithm~\ref{alg:overview},
our method starts by training the feature extractor $f_{\vec\theta_\text{F}}$
by self-supervision on $\cbr{\vec x : (\vec x, y) \in \tilde{\mathcal D}}$.
Then, it learns per-class densities $p_{\vec\theta_y}$
on the features produced by $f_{\vec\theta_\text{F}}$
for each class $y\in\mathcal Y$.
By analyzing the recovered per-class densities, as further 
elaborated in Section~\ref{subsec:target_classes},
we search for disruptively and non-disruptively poisoned target classes.

Once we identify the target classes,
we separate the dataset into three parts.
The clean part $\hat{\mathcal D}_\text{C}$
contains all samples from the clean classes and all
samples from the target classes that receive 
high density of the labeled class
and low density of other classes.
The poisoned part $\hat{\mathcal D}_\text{P}$
contains samples with low density of the class identified as target and 
high density of non-target classes.
The last part $\hat{\mathcal D}_\text{U}$
contains samples with uncertain poisoning status and class membership.
Finally, a discriminative classifier $h_{\vec\theta_{\text C}}$
is trained on $\hat{\mathcal D}_\text{C}$, and fine-tuned on relabeled samples from
$\hat{\mathcal D}_\text{P}'$.

\begin{algorithm}[h!]
\caption{Defense overview}
\begin{algorithmic}[1]
\State Create a feature extractor $f_{\vec\theta_{\text{F}}}$ by self-supervised training on $\tilde{\mathcal D}$. 
\State Learn per-class densities $p_{\vec\theta_{y}}(\vec z)$ of self-supervised 
features $\vec z=f_{\vec\theta_{\text{F}}}(\vec x)$.
\State Identify target classes according to class-level poisoning scores for non-disruptive and disruptive poisoning, $S_\text{ND}^y$ and $S_\text{D}^y$.
\State Assign poisoning scores $\sigma_y(\vec z)$ to all samples from target classes.
\State Based on poisoning scores, split $\tilde{\mathcal D}$ into 
poisoned samples $\hat{\mathcal D}_\text{P}$, 
uncertain samples $\hat{\mathcal D}_\text{U}$, and
clean samples $\hat{D}_\text{C}$.
\State Train a discriminative model $h_{\vec\theta_{\text C}}$ on $\hat{\mathcal D}_\text C$.
\State Produce the relabeled subset $\hat{\mathcal{D}}_\text{P}'$ by relabeling $\hat{\mathcal{D}}_\text{P}$ according to per-class densities.
\State Fine-tune the discriminative model $h_{\vec\theta_{\text C}}$ on $\hat{\mathcal{D}}_\text{P}'$.
\end{algorithmic}
\label{alg:overview}
\end{algorithm}



\subsection{Per-class densities of self-supervised features}

Our defense starts by applying SimCLR~\citep{chen2020simple} to train the self-supervised feature extractor 
$f_{\vec\theta_\text{F}}$ on $\tilde{\mathcal D}$.
Then, we estimate per-class densities of features
$\vec z=f_{\vec\theta_\text{F}}(\vec x)$ 
with lightweight normalizing flows~\cite{kingma2018glow}.
We define a set of normalizing flows with separate parameters $\vec\theta_y$ for each class.
We train per-class densities $p_{\vec\theta_y}(\vec z)$ on the corresponding embedding subsets 
$\mathcal D_\text{F}^y 
    = \cbr{f_{\vec\theta_\text{F}}(\vec x)
        : (\vec x, y') \in \tilde{\mathcal D}, y'=y}$
by maximizing the average log-likelihood:
\begin{equation}
\overline{\mathcal L}(\vec\theta_y, \mathcal D_\text{F}^y) 
= 
\E_{\vec z \in \mathcal D_\text{F}^y}
\log p_{\vec\theta_y} (\vec z) \text.
\end{equation}


After estimating the densities, 
our next objective is to identify the 
classes with poisoned samples and determine
whether the poisoning is disruptive or 
non-disruptive.

\subsection{Identifying target classes}
\label{subsec:target_classes}
We first check for the presence of non-disruptive poisoning by assuming that the poisoned samples 
resemble their source classes
in the self-supervised feature space.
In this case, the generative model of the target class
will assign moderate densities to many foreign samples 
due to learning on triggered samples.
This behaviour will be much less pronounced
in non-target classes.
Consequently, we propose to identify target classes 
according to the average log-density
over all foreign samples:
\begin{align}
    \label{eq:nd_metric}
    \textstyle{
    S_\text{ND}^y = \overline{\mathcal L}\left(\vec\theta_y, \bigcup_{y'\in\mathcal Y \setminus \cbr{y}} \mathcal D_\text{F}^{y'}\right)} \text.
\end{align}
We consider class $y$ as non-disruptively poisoned 
if $S_\text{ND}^y$ exceeds 
the threshold $\beta_\text{ND}$.

Next we check for disruptive poisoning. 
The defining characteristic 
of this type of poisoning
is that the poisoned samples 
are less similar
to all clean samples
than the clean samples 
of different classes
among themselves.
As a consequence, we expect 
the foreign densities in such samples 
to be much lower than the foreign densities 
in the clean samples.
We therefore search for classes 
with a significant number of such outliers. 
We formalize this idea by first defining 
the maximum foreign density score for each sample:
\begin{equation} 
\label{eq:histogram_values}
  v_y(\vec z) = \max_{
    y' \in \mathcal Y\setminus \cbr{y}} 
      p_{\vec\theta_{y'}}(\vec z)
\end{equation}

\begin{figure*}[htb!]
    \centering
    \subfigure[Clean classes]{
    \centering
    \includegraphics[width=0.235\textwidth,trim={0cm 3mm 0 3mm},clip]{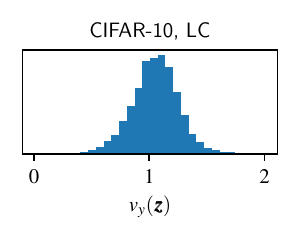}
    \includegraphics[width=0.235\textwidth,trim={0cm 3mm 0 3mm},clip]{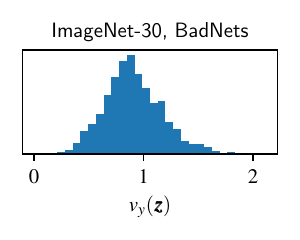}
    }
    \subfigure[Target classes]{
    \centering
    \includegraphics[width=0.235\textwidth,trim={0cm 3mm 0 3mm},clip]{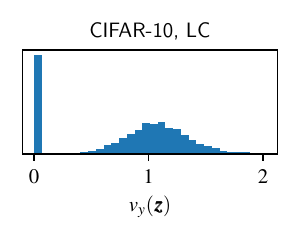}
    \includegraphics[width=0.235\textwidth,trim={0cm 3mm 0 3mm},clip]{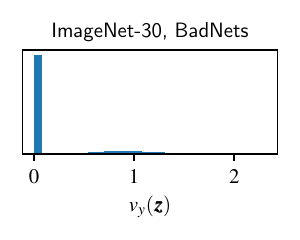}}
    \caption{
    Distributions of the maximum foreign density $v_y(\vec z)$
    of clean and target classes in presence of the Label-Consistent attack~\citep{turner2019label} on CIFAR-10,
    and a strong BadNets attack on ImageNet-30.
    In contrast to clean classes, 
    target classes exhibit
    strong bimodality because poisoned samples
    tend to cluster near $0$. Note that both attacks are disruptive.
    } 
    \label{fig:histograms}
\end{figure*}

We classify a class as disruptively poisoned 
if the fraction of samples 
with low $v$ scores~\eqref{eq:histogram_values}
exceeds the threshold $\beta_\text{D}$.
More precisely, for each class $y$, 
we 
i) compute the set of $v$ scores
 of the corresponding samples 
  $\mathcal V_y = \cbr{
  v_y(\vec z) : \vec z \in \mathcal D_\text{F}^y}$, 
ii) compute a histogram with $30$ bins 
  of equal widths for $\mathcal V_y$ 
  as shown in Figure~\ref{fig:histograms}, 
iii) find the minimum $\mu_y$ of the histogram 
  on the left from the hyperparameter $\lambda$, and
iv) compute the fraction of $\mathcal D_\text{F}^y$ with 
  $v_y(\vec z_\todelete{i})<\mu_y$:
\begin{align}
    \label{eq:d_metric}
  S_\text{D}^y = 
    \frac
     {\envert{\cbr{\vec z \in \mathcal D_\text{F}^y : v_y(\vec z)<\mu_y}}}
     {\envert{\mathcal D_\text{F}^y}}
\end{align}
Finally, we classify a class $y$ 
as disruptively poisoned if $S_\text{D}^y$ 
is less than the threshold $\beta_\text{D}$.
We can interpret $\beta_\text{D}$ 
as the minimum fraction of poisoned samples per class.


\subsection{Identifying poisoned samples}

\begin{figure*}[htb!]
    \centering
    \subfigure[Non-disruptive poisoning attacks]{
    \centering
    \includegraphics[width=0.235\textwidth,trim={0cm 3mm 0 3mm},clip]{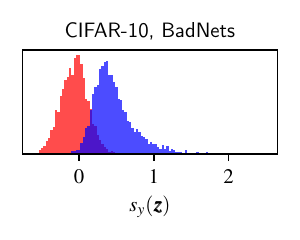}
    \includegraphics[width=0.235\textwidth,trim={0cm 3mm 0 3mm},clip]{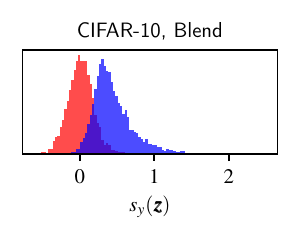}}
    \subfigure[Disruptive poisoning attacks]{ 
    \centering
    \includegraphics[width=0.235\textwidth,trim={0cm 3mm 0 3mm},clip]{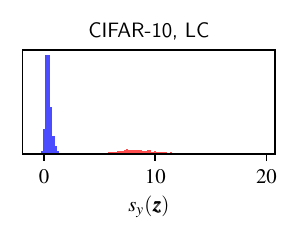}
    \includegraphics[width=0.235\textwidth,trim={0cm 3mm 0 3mm},clip]{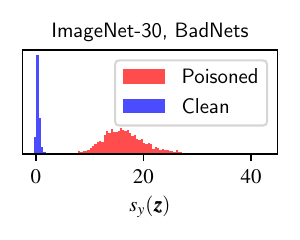}
    }
    \caption{The values of the poisoning score (\ref{eq:score}) for all samples within one target class.
    Clean samples are shown in blue and poisoned samples in red.
    }
    \label{fig:score_values}
\end{figure*}

After identifying the target classes, 
we use the following score to identify poisoned samples:
\begin{equation} \label{eq:score}
    s_y(\vec z) = \frac{p_{\vec\theta_{y}}(\vec z)}{
    \max_{y' \in \mathcal Y\setminus\cbr{y}} p_{\vec\theta_{y'}}(\vec z)} \text.
\end{equation}
The numerator is the density 
of the sample 
with respect to the labeled class.
In the case of non-disruptive poisoning, 
we expect the denominator to be high for poisoned samples 
because they resemble their original class,
and low for clean samples from the target class.
In the case of disruptive poisoning, we expect the densities of disruptively poisoned samples to be very low under all classes but the poisoned one.
Hence, disruptively poisoned samples will score lower
than clean ones, and it will be the opposite in case of non-disruptive poisoning,
as shown in Figure~\ref{fig:score_values}.
Therefore, we define the final poisoning score $\sigma$ so that it is higher for poisoned samples:
\begin{align} 
    \label{eq:sigma}
    \sigma_y(\vec z) = s_y(\vec z)^{1-2{\enbbracket{\text{class $y$ is disruptively poisoned}}}} \text.
\end{align}

\subsection{Filtering and relabeling suspicious samples}

We split the samples 
from identified target classes 
into three parts according 
to the hyperparameters $\alpha_\text{C}, \alpha_\text{P} \in \intoc{0\bidot0.5}$.
We partition the target samples according to the poisoning scores
so that the first $\alpha_\text{P}$ of them are placed in $\hat{\mathcal D}_\text{P}$,
the last $\alpha_\text{C}$ of them in $\hat{\mathcal D}_\text{C}$, while ignoring the 
intermediate. Samples from clean classes are also placed
in $\hat{\mathcal D}_\text{C}$.

Next, we train a classifier on $\hat{\mathcal D}_\text{C}$ using the standard training procedure. 
However, it is possible that $\hat{\mathcal D}_\text{C}$ still contains a small portion of poisoned samples. 
To counteract their influence,  
we proceed to fine-tune the classifier on the relabeled dataset $\hat{\mathcal{D}}_\text{P}'$:
\begin{equation}
    \textstyle{
        \hat{\mathcal D}_\text{P}' = \left\{\left(\vec x, \argmax_{y'\in \mathcal Y\setminus \cbr{y}} p_{\vec\theta_{y'}}(f_{\vec\theta_\text{F}}(\vec x))\right) : (\vec x, y) \in \hat{\mathcal D}_\text{P}\right\} 
    }
\end{equation}
The accuracy of the relabeling process is evaluated in 
Appendix~\ref{app:relabeling_acc}.

\section{Experiments}

\subsection{Experimental setups}

\paragraph{Datasets and models.}
We evaluate our defenses 
on three datasets: 
CIFAR-10 \citep{krizhevsky2009learning}, 
a 30-class subset of ImageNet \citep{deng2009imagenet} (ImageNet-30),
and a 30-class subset of VGGFace2 \citep{cao2018vggface2} (VggFace2-30). 
We use ResNet-18 \citep{he2016deep}
on CIFAR-10 and ImageNet-30,
and DenseNet-121 \citep{huang2017densely} on the VGGFace2-30.
More detailed training setups are given in Appendix~\ref{app:sup_details}.

\paragraph{Attack configurations.}
We consider the following 6 baselines:
BadNets~\citep{gu2019badnets}, blending
attack (Blend)~\cite{chen2017targeted}, 
warping attack (WaNet)~\citep{nguyen2021wanet},
sample-specific triggers (ISSBA)~\citep{li2021invisible},
clean-label attack (LC)~\citep{turner2019label}
and the recent state of the art
based on latent separability
(Adap-Patch and Adap-Blend)~\citep{qi2022revisiting}. 
These baselines
cover visible patch-based attacks (BadNets),
invisible attacks (WaNet and Blend), 
sample-specific attacks (ISSBA) 
and clean-label attacks  (LC). 
We set the target label as $y_\text{T}=0$. 
The poisoning rate is set to $10\%$, 
except for Adap-Patch and Adap-Blend attacks, 
where $1\%$ of the data is poisoned,
and the clean-label attack, where
$2.5\%$ of the data is poisoned.
We omit some attacks 
in ImageNet-30 and VGGFace2-30 experiments
since we were unable to reproduce
the performance from their papers.
Appendix~\ref{att_config} provides detailed 
per-attack configurations.

\paragraph{Defense baselines and configurations.}
We compare our method with four 
state-of-the-art defenses:
Neural attention distillation (NAD)~\citep{li2021neural},
Anti backdoor learning (ABL)~\citep{li2021anti},
Decoupling based defense (DBD)~\citep{huang2022backdoor}
and Backdoor defense 
via adaptive splitting (ASD)~\citep{gao2023backdoor}.
We note that 
NAD and ASD require 
a small subset of clean data
for each class. 
Detailed configurations of our defense and other defenses are provided in Appendix~\ref{app:our_config} and Appendix
\ref{app:defense_configurations}.
Additionally, we validate hyperparameter robustness in Appendix~\ref{app:hyperparameter_val}.

\paragraph{Evaluation metrics.}
We evaluate the two standard metrics of the defense performance, including
the accuracy on the clean test dataset (ACC),
and the attack success rate (ASR) 
that denotes the accuracy 
of recognizing poisoned samples 
as the target label.

\subsection{Performance evaluation}

Table~\ref{tab:main_results} evaluates
effectiveness of our GSSD defense
under state-of-the-art attacks 
and compares it with 
the state-of-the-art.
GSSD consistently achieves 
lower ASR than the alternatives, 
with ASR falling below 1\% in 
the majority of assays.
At the same time, it mantains
consistently high ACC
accross all datasets. 
ASD exhibits slightly higher ACC 
on CIFAR-10 and VGGFace2-30.
However, it often comes at the expense
of significantly higher ASR, even though
ASD requires a small number of clean samples.

\begin{table*}[h]
\newcommand\phbox[1][\phantom{00.0}]{\llap{#1}}

\setlength{\tabcolsep}{1mm}
\footnotesize
\centering
\begin{tabular}{l l rr rr rr rr rr rr}
\toprule
\multicolumn{1}{l}{\multirow{2}{*}[-0.4ex]{Dataset $\downarrow$}}  &
\multicolumn{1}{l}{\multirow{1}{*}{Defense $\rightarrow$}}  &
\multicolumn{2}{c}{No Defense} & 
\multicolumn{2}{c}{NAD*} &
\multicolumn{2}{c}{ABL} &
\multicolumn{2}{c}{DBD} &
\multicolumn{2}{c}{ASD*} &
\multicolumn{2}{c}{GSSD (ours)}  \\
\cmidrule(lr){3-4}
\cmidrule(lr){5-6}
\cmidrule(lr){7-8}
\cmidrule(lr){9-10}
\cmidrule(lr){11-12}
\cmidrule(lr){13-14}
 & Attack $\downarrow$ & ACC & ASR & ACC & ASR & ACC &
ASR & ACC & ASR & ACC & ASR & ACC & ASR \\ 
\midrule
\multirow{8}{*}{CIFAR-10} & BadNets & 94.9 & 100.0 & 88.2 & 4.60 & \textbf{93.8} & \textbf{1.10} &
92.4 & 0.96 & 92.1 & 3.00 & 91.7 & 0.14 \\ 
& Blend & 94.2 & 98.25 & 85.8 & 3.40 & 91.9 & 1.60 &
92.2 & 1.73 & \textbf{93.4} & \textbf{1.00} & 92.2 & 0.77 \\ 
& WaNet & 94.3 & 98.00 & 71.3 & 6.70 & 84.1 &
2.20 & 91.2 & 0.39 & 93.3 & 1.20 & \textbf{93.7} & \textbf{1.35} \\ 
& LC & 94.9 & 99.33 & 86.4 & 9.50 & 86.6 & 1.30 &
89.7 & 0.01 & 93.1 & 0.90 & \textbf{92.8} & \textbf{0.06} \\ 
& ISSBA & 94.5 & 100.0 & 90.7 & 0.64 & 89.2 & 1.20 & 83.2 & 0.53 & 92.4 & 2.13 & \textbf{93.9} & \textbf{0.62} \\ 
& Adap-Patch & 95.2 & 80.9 & 91.1 & 2.96 & 81.9 & 0.00 & 92.9 & 1.77 & 93.6 & 100.0 & \textbf{92.4} &
\textbf{0.23} \\ 
& Adap-Blend & 95.0 & 64.9 & 88.3 & 2.11 & 91.5 & 81.93 & 90.1 & 99.97 & 94.0 & 93.90 & \textbf{92.7} &
\textbf{0.22} \\ 
\cmidrule{2-14}
& Average & - & - & 85.6 & 4.27 & 88.4 & 12.76 &
90.2 & 15.05 & 93.1 & 28.87 & \textbf{92.6} & \textbf{0.48}  \\ 
\midrule
\multirow{4}{*}{ImageNet-30} & BadNets & 95.3 & 99.98 & 92.7 & 0.42 & 94.3 & 0.24 & 91.2 &
0.54 & 90.7 & 9.72 & \textbf{94.8} & \textbf{0.00} \\ 
& Blend & 93.7 & 99.93 & 90.0 & 0.51 & 93.1 & 0.14
& 90.3 & 0.58 & 89.9 & 2.07 & \textbf{93.5} & \textbf{0.45} \\ 
& WaNet & 93.5 & 100.0 & 90.7 & 0.56 & 92.0 & 1.33 &
90.5 & 0.48 & 88.8 & 2.89 & \textbf{93.4} & \textbf{1.33} \\ 
\cmidrule{2-14}
& Average & - & - & 91.1 & 0.50 & 93.1 & 1.71 &
90.7 & 0.53 & 89.8 & 4.89 & \textbf{93.9} & \textbf{0.59}   \\ 
\midrule
\multirow{4}{*}{VGGFace2-30} & BadNets & 93.2 & 100.0 & 56.1 & 6.50 & 93.9 & 2.67 & 90.3 & 0.00 
& 96.7 & 98.12  & \textbf{95.2} & \textbf{0.09} \\ 
& Blend & 92.8 & 99.95 & 50.8 & 7.30 & 93.4 & 5.40
& 90.2 & 0.04 & \textbf{96.0} & \textbf{0.18} & 94.2 & 0.42 \\ 
& WaNet & 93.7 & 99.60 & 50.4 & 4.20 & 93.2 & 1.48 & 
87.2 & 0.00 & 96.9 & 96.51 & \textbf{94.9} & \textbf{0.14} \\ 
\cmidrule{2-14}
& Average & - & - & 52.4 & 6.00 & 93.5 & 3.18 &
89.2 & 0.01 & 96.5 & 64.94 & \textbf{94.8} & \textbf{0.22}   \\ 
\bottomrule
\end{tabular}
\caption{Comparisons of the proposed GSSD defense with 4
baselines on CIFAR-10, ImageNet-30 and VGGFace2-30 datasets.
We mark defenses requiring extra clean data with~*. We report 
the clean test accuracy as ACC [\%] and the attack
success rate ASR [\%]. The defense that exhibits the highest 
ACC $-$ ASR is marked in \textbf{bold}.}
\label{tab:main_results}
\end{table*}

We highlight strong defense performance of GSSD 
under Adap-Patch and Adap-Blend attacks.
Despite the aim of these attacks 
to suppress the latent separation 
between poisoned and clean samples, 
self-supervised representations
of poisoned samples 
are still in high-density regions 
of their original class.
Conversely, when assessing 
ABL, DBD, and ASD defenses 
against Adap-Blend, 
ASR increases
compared to no defense.




\paragraph{Robustness to different poisoning rates }
Table~\ref{tab:poisoning_rates_table} validates the resistance 
against three representative poisoned label attacks with different poisoning rates.
GSSD manages to reduce ASR to below or 
around 1\% in all cases, while retaining high accuracy on benign inputs.

\begin{table*}[htb!]
    \small
    \centering
    \begin{tabular}{r l rr rr rr}
    \toprule
     \multirow{2}{*}[-0.4ex]{Poisoning rate $\downarrow$} &
     \multicolumn{1}{l}{Attack $\rightarrow$} &
     \multicolumn{2}{c}{BadNets} &
     \multicolumn{2}{c}{Blend} 
     & \multicolumn{2}{c}{WaNet} \\
     \cmidrule(lr){3-4}
     \cmidrule(lr){5-6}
     \cmidrule(lr){7-8}
     & Defense $\downarrow$  & ACC & ASR & ACC & ASR & ACC & ASR \\
     \midrule
     \multirow{2}{*}{1\%} & No Defense & 95.2 & 99.96 & 95.2 & 94.52 & 94.7 & 60.05 \\ 
     & GSSD & 91.2 & 0.04 & 91.7 & 0.05 & 91.1 & 0.01 \\ 
     \midrule
     \multirow{2}{*}{5\%} & No Defense & 94.5 & 100.00 & 94.7 & 99.34 & 94.4 & 95.70 \\ 
     & GSSD & 93.5 & 0.47 & 93.5 & 1.30 & 93.4 & 0.52 \\ 
     \midrule
     \multirow{2}{*}{10\%} & No Defense & 95.0 & 100.00 & 94.6 & 99.69 & 94.5 & 99.00 \\ 
     & GSSD & 91.7 & 0.14 & 92.2 & 0.77 & 93.7 & 1.35 \\ 
     \midrule
     \multirow{2}{*}{15\%} & No Defense & 94.5 & 100.00 & 94.3 & 99.92 & 94.2 & 99.71 \\ 
     & GSSD & 92.8 & 0.44 & 92.1 & 0.94 & 93.0 & 0.83 \\ 
     \midrule
     \multirow{2}{*}{20\%} & No Defense & 94.5 & 100.00 & 94.4 & 99.90  & 94.1& 98.83 \\ 
     & GSSD & 91.6 & 1.20 & 89.6 & 1.90 & 92.5& 1.00 \\ 
     \bottomrule
    \end{tabular}
    \caption{Robustness of GSSD to different poisoning rates on 
    CIFAR-10.} 
    \label{tab:poisoning_rates_table}
\end{table*}

\paragraph{Robustness to adaptive attacks.}
We analyze the effect of a
potential adaptive attack in Appendix~\ref{sec:adaptive},
finding that GSSD successfully defends against it.

\paragraph{Time complexity.}
Appendix~\ref{app:time} presents time complexity measurements.
Although it ranks second to ABL, GSSD outperforms DBD and ASD significantly 
in terms of time efficiency.

\paragraph{Limitations.} 
Even though our defense has proven successful against 
all considered state-of-the-art
attacks through thorough empirical testing, it is important 
to note that it lacks theoretical guarantees. 
We have no precise characterization of
what kinds of attacks it will be resilient against given the properties of the dataset.
A weakness of detecting target classes as proposed here is that, 
if the target class detection step in 
Equation~\eqref{eq:nd_metric} fails, then the entire defense fails.
We have observed such behavior in attack that can succeed
with very low poisoning rate. 
Current state of the art~\cite{li2021anti, gao2023backdoor} also 
fails in such scenarios. 
Further discussion is
provided in Appendix~\ref{sec:poisoning_rates}. 
As introduced in section~\ref{subsec:formulation}, we are working 
under the assumption that all poisoned
samples share the same target label, a scenario commonly referred to as 
\textit{all-to-one} poisoning. There are other types of poisoning in literature,
such as \textit{all-to-all} poisoning, which are beyond the scope or our work.

\subsection{Ablation studies}

\paragraph{The importance of self-supervision.} 
Table~\ref{tab:features_comparison} validates the choice of the pre-trained feature extractor.
We compare our choice of SimCLR 
with supervised training and CLIP~\cite{radford2021learning}.
We can see that SimCLR and CLIP 
deliver similar overall performance, 
while greatly outperforming 
the supervised representations.
This improvement occurs
since self-supervised learning 
is less affected by triggers 
and not affected by target labels. 
Interestingly, our defense 
still works well and 
maintains high accuracy 
against the WaNet attack 
even with supervised representations.
%
Additionally, we notice that CLIP performs quite well. 
This suggests that the computationally expensive pre-training on the poisoned dataset may not be necessary. 
Instead, using pre-trained feature extractors like CLIP can be effective.

\begin{table}[htb!]
    \small
    \centering
    \begin{tabular}{l rr rr rr} 
    \toprule
     \multicolumn{1}{l}{Attack $\rightarrow$} & \multicolumn{2}{c}{BadNets} & 
     \multicolumn{2}{c}{Blend} &
     \multicolumn{2}{c}{WaNet} \\
     \cmidrule(lr){2-3}
     \cmidrule(lr){4-5}
     \cmidrule(lr){6-7}
     Feature extractor $\downarrow$ & ACC & ASR & ACC & ASR & ACC & ASR \\ 
     \midrule
     RN-18 supervised  & 75.1 & 6.00 & 76.7 & 76.40 & 92.4 & 0.42 \\ 
     RN-50 CLIP & 91.1 & 0.10 & \textbf{93.0} & \textbf{0.88} & \textbf{94.0} & \textbf{0.81} \\ 
     RN-18 self-sup (SimCLR) & \textbf{91.7} & \textbf{0.14} & 92.2 & 0.77 & 93.7 & 1.35\\ 
     \bottomrule
    \end{tabular}
    \caption{Comparison of different feature extractors
    on CIFAR-10.} 
    \label{tab:features_comparison}
\end{table}

\paragraph{Effect of generative classification.}
Appendix~\ref{app:genvsdis} 
highlights the strenghts
of the generative classifier compared to the discriminative classifier trained on the same features.
The discriminative classifier fails at detection of target classes, but performs well at filtering and relabeling given knowledge of the target class.

\paragraph{The impact of fine-tuning on relabeled data.}
Table~\ref{tab:finetuning} validates the impact of our method
during standard discriminative training
and subsequent fine-tuning.
Top two rows show that
fine-tuning with relabeled data
breaks the backdoor
even after standard training
on poisoned data.
The remaining rows indicate that
we obtain the best ACC$-$ASR performance
with training
on the filtered clean subset $\hat{\mathcal{D}}_\text C$
followed by fine-tuning on
the relabeled subset $\hat{\mathcal{D}}_\text{P}'$.

\begin{table}[!ht]
    \centering
    \small
    \begin{tabular}{l c rr rr rr rr}
    \toprule
        \multirow{2}{*}{Training data} & \multirow{2}{*}{\shortstack[c]{Fine-\\tuning}} & \multicolumn{2}{c}{Badnets} &
        \multicolumn{2}{c}{Blend } &
        \multicolumn{2}{c}{Wanet} &
        \multicolumn{2}{c}{Adap-Patch} \\ 
        \cmidrule(lr){3-4}
        \cmidrule(lr){5-6}
        \cmidrule(lr){7-8}
        \cmidrule(lr){9-10}
        & & ACC & ASR & ACC & ASR & ACC & ASR & ACC & ASR \\
        \midrule
        Original ($\tilde{\mathcal{D}}$) & \textcolor{lightgray}{\xmark} & 94.5 & 100.0 & 94.2 & 98.32 & 94.5 & 99.12 & 95.2 & 80.95  \\ 
        Original ($\tilde{\mathcal{D}}$) & \cmark & 93.6 & 1.48 & 92.0 & 1.30 & 94.0 & 3.47 & 94.3 & 8.01  \\ 
        Cleansed ($\hat{\mathcal{D}}_\text C$) & \textcolor{lightgray}{\xmark} & 93.8 & 95.92 & 94.0 & 51.00 & 93.6 & 5.30 & 93.4 & 0.18  \\ 
        Cleansed ($\hat{\mathcal{D}}_\text C \cup \hat{\mathcal{D}}_\text{P}'$) & \textcolor{lightgray}{\xmark} & 93.8 & 3.69 & 93.5 & 10.47 & 93.5 & 2.58 & 92.2 & 0.10  \\
        Cleansed ($\hat{\mathcal{D}}_\text C$) & \cmark & 91.7 & 0.14 & 92.2 & 0.77 & 93.7 & 1.35 & 92.4 & 0.23  \\ 
        \bottomrule
    \end{tabular}
    \caption{Validation of the impact of fine-tuning with the relabeled subset $\hat{\mathcal{D}}_\text{P}'$. 
    The original dataset is the poisoned dataset $\tilde{\mathcal{D}}$, the input to our method. 
    The filtered clean subset $\hat{\mathcal{D}}_\text C$ and the relabeled poisoned subset $\hat{\mathcal{D}}_\text{P}'$ are produced by our method.}
    \label{tab:finetuning}
\end{table}


\section{Conclusion}


We have presented a novel analysis 
of the effects of backdoor attacks 
onto self-supervised image representations.
The results inspired us 
to propose a novel backdoor defense
that allows to detect 
target classes and samples.
Extensive evaluation against 
the state-of-the-art 
reveals competitive performance.
In particular, we note
extremely effective ASR reduction 
in presence of latent separability attacks
Adap-Patch and Adap-Blend.
We hope that our method 
can contribute as a tool
for increasing the robustness 
of deep learning applications.
Suitable directions for future work include 
circumventing the detection of target classes
and extending the applicability 
to other kinds of poisoning.

\paragraph{Acknowledgments.}
This work has been co-funded by the European Defence Fund grant EICACS and Croatian Science Foundation grant IP-2020-02-5851 ADEPT.

\bibliography{egbib}
\appendix

\numberwithin{equation}{section}
\numberwithin{table}{section}
\numberwithin{figure}{section}

\section{Normalizing flows}
\label{app:norm_flows}
A normalizing flow is a bijective mapping $g_{\vec\theta_\text{NF}}$ that transforms the input $\rvec z$ with a complex distribution into an output $\rvec u$ with a fixed simple distribution, usually an isotropic Gaussian with a zero mean and unit variance: $g_{\vec\theta_\text{NF}}(\rvec z) = \rvec u \sim \mathcal{N}(\cvec 0_d, \cvec I_d)$, where $d$ is the dimension of the input.
The density of the inputs can be computed by applying change of variables:
\begin{equation} \label{eq:change_of_variables}
p(\vec z) = p(\vec u) \left|\det \frac{\partial \vec u}{\partial \vec z}\right|
\end{equation}
A normalizing flow is usually implemented as a sequence of simpler invertible mappings with learnable parameters, such as affine coupling layers~\cite{dinh2016density}.

\section{Implementation details}  
\label{app:sup_details}

\subsection{Datasets and models}
We show the details for each dataset used in our evaluations
in Table~\ref{tab:datasets_dnns}. 
Following previous work \cite{huang2022backdoor,gao2023backdoor}, We use subsets of 30 classes from ImageNet~\cite{deng2009imagenet}
and VGGFace2~\cite{cao2018vggface2}, primarily to address computational time and cost constraints.
We have chosen the subsets randomly because the subsets of previous works are not publicly available.
We call the subsets ImageNet-30 and VggFace2-30.
The classes selected for ImageNet-30 are: \textit{acorn, airliner, ambulance, american alligator banjo, barn, bikini, digital clock, dragonfly, dumbbell, forklift, goblet, grand piano, hotdog, hourglass, manhole cover, mosque, nail, parking meter, pillow, revolver, rotary dial telephone, schooner, snowmobile, soccer ball, stringray, strawberry, tank, toaster, volcano}.
The classes selected for VGGFace2-30 are: \textit{557, 788, 1514, 2162, 2467, 3334, 3676, 4908, 5491, 5863, 6248, 7138, 7305, 7620, 8316 591, 1480, 2035, 2251, 2933, 3416, 4215, 5318, 5640, 5891, 7084, 7222, 7489, 8144, 8568}.

\begin{table*}[h]
\centering
\begin{tabular}{llrrrl}
\toprule
\multicolumn{1}{l}{\multirow{2}{*}{Dataset}} & \multicolumn{1}{c}{\multirow{2}{*}{Input size}}  & \multicolumn{1}{r}{\multirow{2}{*}{\# Classes}} & \multicolumn{1}{r}{\# Training}  & \multicolumn{1}{r}{\# Testing} & \multirow{2}{*}{Model}\\
 &  &  &  images &  images & \\
\midrule
CIFAR-10 & 3 $\times$ 32 $\times$ 32 & 10 & 50000 & 10000 & ResNet-18  \\
ImageNet-30 & 3 $\times$ 224 $\times$ 224 & 30 & 39000 & 3000 & ResNet-18 \\
VGGFace2-30 & 3 $\times$ 224 $\times$ 224 & 30 & 9000 & 2250 & DenseNet-121 \\
\bottomrule
\end{tabular}
\caption{Summary of datasets and models used in our experiments.}
\label{tab:datasets_dnns}
\end{table*}

\subsection{Standard supervised training setups}

Our experiments involve
the ResNet-18 backbone~\cite{he2016deep} with 
the standard stem block depending on the dataset.
For ImageNet, the stem block consists of a $7\times 7$ convolution with stride $2$ followed by batchnorm, ReLU and $3\times 3$ average pooling with stride $2$.
For CIFAR-10, the stem block is a single $3\times 3$ convolution with stride $1$.

We perform supervised training on CIFAR-10~\cite{krizhevsky2009learning} 
for 200 epochs with a batch size of 128. 
We use SGD with momentum set to 0.9 and weight decay to 0.0005.
Following~\cite{huang2022backdoor, gao2023backdoor}, the initial learning rate is $0.1$,
and we divide it by 10 at epochs 100 and 150.
We perform random resized crop, random horizontal flip as data augmentations and standardize the 
inputs~\cite{huang2022backdoor}.

On ImageNet-30 and VGGFace2-30 we train for 90 epochs. 
All images are resized to 
$224 \times 224$ before the trigger injection.
The other hyperparameters are same as in CIFAR-10 training.

\section{Attack configurations}
\label{att_config}

\paragraph{BadNets}
To perform BadNets attacks, we follow the configurations of 
\cite{gu2019badnets, huang2022backdoor, gao2023backdoor}. On CIFAR-10, the 
trigger pattern is a $2 \times 2$
square in the upper left corner of the image. 
On ImageNet-30 and VGGFace2-30, we opt for a
$32 \times 32$ Apple logo.

\paragraph{Blend}
Following~\cite{chen2017targeted, huang2022backdoor, gao2023backdoor}, 
we use "Hello Kitty" pattern on CIFAR-10 and random noise patterns on ImageNet-30
and VGGFace2-30. Blending ratio on all datasets is set to 0.1.

\paragraph{WaNet}
Although WaNet~\cite{nguyen2021wanet} belongs to the training time attacks, we follow
\cite{huang2022backdoor, gao2023backdoor} to use the warping-based operation to directly generate
the trigger pattern. The operation hyperparameters are the same as in~\cite{gao2023backdoor}.

\paragraph{Label Consistent}
Following~\cite{turner2019label}, we use projected gradient descent 
\cite{madry2017towards}
to generate the adversarial perturbations within $L^{\infty}$ ball.
Maximum magnitude $\eta$ is set to 16, step size to 1.5 and perturbation steps 
to 30. Trigger pattern is $3 \times 3$ grid pattern in each corner of the image

\paragraph{ISSBA}
We replicate the ISSBA~\cite{li2021invisible} attack
by training the encoder model for 20 epochs with secret size 20.
We then leverage the pre-trained encoder to generate the poisoned dataset.

\paragraph{Adap-Patch and Adap-Blend}
To replicate these attacks, we search for the combination of
cover and poison rate giving the best ASR, while keeping in mind that 
those rates should not be too high for 
attack to remain stealthy, as stated in~\cite{qi2022revisiting}.
We set poisoning and cover rate both to $0.01$.
Trigger patterns used 
are the same as in~\cite{qi2022revisiting}.

\section{Defense configurations}
\label{app:defense_configurations}

\paragraph{NAD}
We implement NAD based on the open source
code of the BackdoorBox
library\footnote{\url{https://github.com/THUYimingLi/BackdoorBox}}.
We find it~\cite{li2021neural} to be sensitive to its hyperparameter 
$\beta$. Therefore, for every attack, we perform a hyperparameter search for the best
results among values $\beta \in \{500, 1000, 1500, 2000, 5000\}$.

\paragraph{ABL}
To reproduce ABL experiments, we refer to BackdoorBox.
We first poison the model
for 20 epochs, followed by backdoor isolation which takes 
70 epochs. 
Lastly, we unlearn the backdoor for 5 epochs on CIFAR-10 and ImageNet-30,
and for 20 epochs on VGGFace2-30.
We search for the value of the hyperparameter $\gamma\in \{0, 0.2, 0.4\}$ that gives the best ASR.

\paragraph{DBD}
In order to reproduce DBD~\cite{huang2022backdoor}, we use the official implementation\footnote{\url{https://github.com/SCLBD/DBD}}. We follow all configurations as described
in~\cite{huang2022backdoor}.

\paragraph{ASD}
By following the official implementation\footnote{\url{https://github.com/KuofengGao/ASD}},
we reproduce the ASD~\cite{gao2023backdoor}. 
We follow all defense configurations
from~\cite{gao2023backdoor}

\section{GSSD}
\label{app:our_config}

The self-supervised and the supervised stage of GSSD involve
the ResNet-18 backbone~\cite{he2016deep} described in Appendix~\ref{app:sup_details}.

We perform SimCLR~\cite{chen2020simple} pre-training for 100 epochs on batches of 256 images. 
We use Adam with $(\beta_1,\beta_2)=(0.9, 0.99)$, 
and a fixed learning rate of $3\cdot10^{-4}$.
We perform random resized crop, random horizontal flip, color jitter, grayscale as data augmentations
and standardize the inputs~\cite{chen2020simple}. 
The code 
implementation\footnote{\url{https://github.com/Spijkervet/SimCLR}}
of SimCLR that we use omits
random Gaussian blurring compared to the original paper.

Our per-class normalizing flows consist of two steps with actnorm~\cite{kingma2018glow} and affine coupling~\cite{dinh2016density}.
Each coupling module computes the affine parameters with a pair of
ReLU activated fully-connected layers.
We train each normalizing flow for 50 epochs with batch size 16, use Adam optimizer with 
$(\beta_1,\beta_2)=(0.9, 0.99)$ 
and a fixed learning rate $\delta=10^{-3}$.

After the standard supervised trainingwith hyperparameters described in Appendix~\ref{app:sup_details}, we
fine-tune the classifier for 2 epochs using the learning
rate of $10^{-4}$.
We set $\beta_\text{ND}=0.6$, 
$\beta_\text{D}=0.05$, 
$\lambda=0.75$, $\alpha_\text{C}=0.3$ and 
$\alpha_\text{P}=0.15$
according to early validation experiments. 
The defense against disruptive attack uses $\alpha_C=0.15$.
Appendix~\ref{app:hyperparameter_val} provides extensive
hyperparameter validations.

\section{Attack impact on self-supervised embeddings}
\label{app:attack_impact}

\begin{figure}[htb]
    \centering
    \includegraphics[width=.5\textwidth]{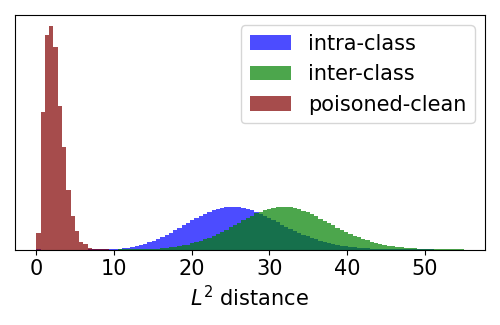}
    \caption{Histogram of $L^2$ distances between self-supervised embeddings for
    BadNets attack on CIFAR-10 dataset. 
    Distances between clean and poisoned
    versions of the same example are colored in brown. Blue denotes distances between samples of the 
    same classes (intra-class), while green
    marks distances between samples from different classes (inter-class).
    }
    \label{fig:l2_distances}
\end{figure}

Figure~\ref{fig:l2_distances} shows how adding triggers into images affects 
self-supervised embeddings in non-disruptive attacks.
Concretely, we measure the $L^2$ distance between embeddings of the same image
before and after trigger addition. 
We compare these distances with those between examples of the same label, as well 
as with distances between examples of different labels. We conclude that the impact
of the trigger injection is minimal. The poisoned examples will be much more similar 
to clean examples from their original class than to the target class examples.


\section{Time complexity} 
\label{app:time}

Table~\ref{tab:time} compares the runtimes of different defenses.
We utilize the official implementations 
provided by the authors (see Appendix~\ref{app:defense_configurations})
to measure the runtime of each method.
The experiments on 
CIFAR-10 were conducted on Nvidia RTX 2080 Ti, and 
the experiments on ImageNet-30 were conducted on 
Nvidia RTX A4500 due to greater memory requirements. 
Note that we were unable to achieve full GPU utilization on our machines.
Therefore, we also provide the measurements from~\cite{gao2023backdoor}, that had better GPU utilization.

ABL requires the least amount of time on both
CIFAR-10 and ImageNet-30 datasets. 
However, it is quite sensitive to its hyperparameter, which varies 
inconsistently across different datasets.
Further adjustment of this hyperparameter adds complexity 
to the defense process, resulting in increased time requirements.
Our method is more efficient than DBD and ASD. 
The computationally most expensive part of DBD and ASD is the mixmatch semi-supervised stage, while in case of GSSD it is the self-supervised stage.
By relabeling suspicious examples and employing standard supervised learning, our method avoids the time consuming semi-supervised learning.
In case of GSSD, most of the time is spent on self-supervised training, as shown in the breakdown over stages in Table~\ref{tab:time_component}.
The self-supervised training stage might be replaced with an off-the-shelf 
feature extractor, as shown in Table~\ref{tab:features_comparison}.
However, the assumption of obtaining 
a clean pre-trained feature extractor, such as CLIP, may face challenges, 
as recent research shows how similar models can be backdoored~\cite{carlini2021poisoning, carlini2023poisoning}.

\begin{table}[!ht]
    \centering
    \begin{tabular}{ll rrrr}
        \toprule
         GPU & Dataset & ABL & DBD & ASD & GSSD \\ 
         \midrule
         RTX 2080 Ti & CIFAR-10 & \textbf{4825} & 19282 & 13202 & \underline{6013} \\
         RTX 2080 Ti \cite{gao2023backdoor} & CIFAR-10 & \textbf{3200} & *45850 & 9990 & -- \\
         RTX A4500 & ImageNet-30 & \textbf{4154} & 37007 & 22557 & \underline{16217} \\
         Tesla V100 \cite{gao2023backdoor} & ImageNet-30  & \textbf{3855} & *223100 & 27030 & -- \\
        \bottomrule
    \end{tabular}
    \caption{Runtime [$\mathrm s$] of defenses against the BadNets attack.
    Rows with reference to~\cite{gao2023backdoor} correspond to measurements copied from~\cite{gao2023backdoor}.
    The symbol * denotes that the measurements from~\cite{gao2023backdoor} used $1000$ as the number of epochs for self-supervised training, while we reduced it to $100$ based on the observation of no substantial effect on the performance.}
    \label{tab:time}
\end{table}

\begin{table}[!ht]
    \centering
    \begin{tabular}{l rrrr}
        \toprule
         Dataset & SimCLR & Normalizing flows & Retraining & Total \\ 
         \midrule
         CIFAR-10 & 2172 & 896 & 2945 & 6013 \\
         ImageNet-30 & 11940 & 946 & 3331 & 16217 \\
        \bottomrule
    \end{tabular}
    \caption{Runtime [$\mathrm{s}$] of each component of GSSD. SimCLR denotes self-supervised
    training in the first stage of GSSD. 
    Normalizing flows denote the training of 
    and filtering and relabeling with the generative classifier. 
    Retraining marks training on the filtered subset and fine-tuning on the relabeled subset.}
    \label{tab:time_component}
\end{table}

\section{Resistance to low poisoning rates}
\label{sec:poisoning_rates}

We have observed that GSSD fails to detect the target class in the case of an
extremely low poisoning rate, such as 0.1\%. BadNets is the only attack that succeeds 
in such a scenario. We have noticed that other state-of-the-art methods also struggle against 
that attack, as shown in Table~\ref{tab:0-1_poisoning_rate}.

\begin{table*}[htb!]
    \centering
    \footnotesize
    \begin{tabular}{r rr rr rr rr rr}
    \toprule
     Defense $\rightarrow$ &
     \multicolumn{2}{c}{No defense} &
     \multicolumn{2}{c}{ABL} &
     \multicolumn{2}{c}{DBD} &
     \multicolumn{2}{c}{ASD} 
     & \multicolumn{2}{c}{GSSD} \\
     \cmidrule(lr){2-3}
     \cmidrule(lr){4-5}
     \cmidrule(lr){6-7}
     \cmidrule(lr){8-9}
     \cmidrule(lr){10-11}
      Poisoning rate $\downarrow$  & ACC & ASR & ACC & ASR & ACC & ASR & ACC & ASR & ACC & ASR \\
     \midrule
     0.1\% & 95.2 & 99.06 & 84.3 & 99.51 & \textbf{91.5} & \textbf{2.45} & 93.1 & 93.43 & 95.2 & 99.06 \\ 
    \bottomrule
    \end{tabular}
    \caption{Performance of state-of-the-art defenses against BadNets attack with 0.1\%
    poisoning rate on CIFAR-10.} 
    \label{tab:0-1_poisoning_rate}
\end{table*}

\section{Resistance to potential adaptive attacks}
\label{sec:adaptive}

Adaptive attacks are crafted by attackers who have knowledge of potential defense methods.
In our case, the attacker could try to fool a surrogate self-supervised
model by increasing similarity between poisoned samples and the clean 
samples of the target class.
One way to achieve this is to search for a trigger that minimizes some distance $d$ between poisoned samples and clean samples of the target class. 

Let $t_{\vec\tau}\colon \mathcal X\to \mathcal X$ denote the triggering function that applies a blending trigger pattern $\vec\tau\in \mathcal X = \intcc{0, 1}^{H\times W\times 3}$ with weight $b\in \intoo{0, 1}$ to an example $\vec x$ as follows:
\begin{align}
    t_{\vec\tau}(\vec x) = (1-b) \vec x + b\vec\tau \text.
\end{align}
We optimize the trigger pattern $\vec\tau$ on a surrogate self-supervised model trained on images from the benign dataset $\mathcal D$.
Let $\mathcal D_\text{p}\subset \mathcal D$ be the subset of training examples to be triggered.
The clean examples $\mathcal{D}_\text{C} = \mathcal{D} \setminus \mathcal D_\text{p}$ are the 
rest of the dataset.
Formally, we
aim to solve
\begin{equation}
\label{eq:adaptive}
    \min_{\vec\tau} \sum_{(\vec x, y) \in \mathcal{D}_\text{p}} d(\bar{\vec z}_{y_{\text T}}, f_{\vec\theta_\text F}(t_{\vec\tau}(\vec x) )) \text,
\end{equation}
where 
$ \bar {\vec z}_{y_{\text T}} \triangleq \E_{(\vec x, y) \in 
D_\text C, y = y_T} f_{\vec\theta_\text F} (\vec x)$ 
is the average embedding of clean samples of the target class.
We set $b=0.2$ and set the poisoning rate to $10\%$.

To maximize the influence of the trigger $\vec\tau$ in the latent space, we size it to be the same
as that of the original image.
This makes the attack less stealthy, but our goal is to test the resilience of our defense against the most potent attack possible.
In this attack scenario, we assume the attacker has access to 
the benign training dataset, self-supervised model structure and optimization objective.

We perform experiments on CIFAR-10 using the Adam optimizer with learning rate set to $0.1$. 
The attack results in a successful backdoor with $\mathrm{ACC}=94.7\%$ and $\mathrm{ASR}=100\%$. 
However, GSSD successfully detects the poisoned samples and erases the backdoor during retraining process. 
The final result is $\mathrm{ACC}=93.6\%$, $\mathrm{ASR}=0.27\%$.
It classifies this attack as disruptive and
filters out all poisoned samples.
We hypothesise that the attacker faces a compromise: a stronger trigger is more likely to 
to minimize the distance in Eq.~\eqref{eq:adaptive}, but also more likely to make examples with triggers more similar to each other, thereby raising the risk of the poisoning being disruptive.

\section{Generative vs Discriminative classifier} 
\label{app:genvsdis}
To validate the choice of the generative classifier in our method, we plug in
the discriminative classifier in its place. We use a simple
model consisting of two linear transformations with ReLU in between and 
optimize it using standard cross entropy loss. The discriminative classifier
fails at detecting target classes using Equations~\eqref{eq:nd_metric} and~\eqref{eq:d_metric}. 
For the rest of this ablation, we assume that the defender knows the target class $y_\text T$.
We utilize the predictions from a discriminative classifier
to compute $\sigma$ from Equation~\eqref{eq:sigma}, which is then employed to 
perform filtering and relabeling. 
Table~\ref{tab:classifiers_comparison} compares the performance of a model trained
on such data against the original version of our method with the generative classifier.
Despite the slight reduction in accuracy on clean data when using 
the discriminative classifier in our method, it still produces satisfactory results.

\begin{table}[htb!]
    \centering
    \begin{tabular}{l rr rr rr} 
    \toprule
     Attack $\rightarrow$ & \multicolumn{2}{c}{BadNets} & 
     \multicolumn{2}{c}{Blend} &
     \multicolumn{2}{c}{WaNet} \\
     \cmidrule(lr){2-3}
     \cmidrule(lr){4-5}
     \cmidrule(lr){6-7}
     Classifier $\downarrow$ & ACC & ASR & ACC & ASR & ACC & ASR \\ 
     \midrule
     Discriminative & 90.5 & 0.05 & 91.7 & 0.29 & 92.9 & 1.30 \\ 
     Generative & \textbf{91.7} & \textbf{0.23} & \textbf{92.2} 
     & \textbf{0.77} & \textbf{93.7} & \textbf{1.35} \\ 
     \bottomrule
    \end{tabular}
    \caption{Comparison of different classifiers
    on CIFAR-10.} 
    \label{tab:classifiers_comparison}
\end{table}


\section{Hyperparameter validation}
\label{app:hyperparameter_val}
We provide validation for hyperparameters $\alpha_\text C$ and 
$\alpha_\text P$
in 
Table~\ref{tab:alpha},  
for $\beta_\text D$ in Tables~\ref{tab:beta_d_1}, 
\ref{tab:beta_d_2}, 
for $\lambda$ in~\ref{tab:lambda_1},~\ref{tab:lambda_2}, and
for $\beta_\text{ND}$ in~\ref{tab:beta_nd_1},~\ref{tab:beta_nd_2}.

\begin{table*}[htb!]
    \centering
    \begin{tabular}{rr rr rr rr rr}
    \toprule
     \multicolumn{2}{r}{Attack $\rightarrow$} & \multicolumn{2}{c}{BadNets} & 
     \multicolumn{2}{c}{Blend} &
     \multicolumn{2}{c}{Wanet} &
     \multicolumn{2}{c}{Adap-Blend} \\
     \cmidrule(lr){3-4}
     \cmidrule(lr){5-6}
     \cmidrule(lr){7-8}
     \cmidrule(lr){9-10}
     $\alpha_\text{C} \downarrow$ & $\alpha_\text{P} \downarrow$ & ACC & ASR & ACC & ASR & ACC & ASR & ACC & ASR  \\
     \midrule
     0.15 & 0.15 & 92.2 & 0.20 & 91.4 & 0.75 & 92.5 & 0.60 & 90.0 & 0.00 \\ 
     0.3 & 0.3 & 93.0 & 0.34 & 93.0 & 0.80 & 93.4 & 0.90 & 89.2 & 0.00 \\ 
     0.4 & 0.4 & 92.9 & 0.30 & 92.3 & 0.75 & 92.5 & 0.47 & 87.4 & 0.00 \\ 
     0.4 & 0.15 & 92.6 & 0.40 & 91.1 & 1.10 & 93.3 & 1.20 & 91.5 & 0.02 \\ 
     0.3 & 0.15 & 92.4 & 0.20 & 92.6 & 0.84 & 93.7 & 1.20 & 91.3 & 0.01 \\ 
     0.4 & 0.3 & 93.4 & 0.51 & 92.5 & 1.03 & 93.8 & 1.20 & 89.1 & 0.00 \\ 
     \bottomrule
    \end{tabular}
    \caption{Results of our defense for BadNets attack on CIFAR-10 for different values of hyperparameters $\alpha_\text{C}$ and $\alpha_\text{P}$. } 
    \label{tab:alpha}
\end{table*}


\begin{table*}[htb!]
    \centering
    \begin{tabular}{r c c c c}
    \toprule
     $\beta_\text D \rightarrow$ &
     \multirow{2}{*}{0.01} & 
     \multirow{2}{*}{0.05} &
     \multirow{2}{*}{0.10} &
     \multirow{2}{*}{0.20} \\
     Poisoning rate $\downarrow$ & & & & \\
     \midrule
     0.65\% & \textit{airplane} & \textit{airplane} & \textcolor{red}{none} & \textcolor{red}{none} \\ 
     2.50\% & \textit{airplane} & \textit{airplane} & \textit{airplane} & \textit{airplane} \\ 
    \bottomrule
    \end{tabular}
    \caption{Results of target classes detection for LC attack on CIFAR-10 dataset for different 
    values of hyperparameter $\beta_\text D$. The true target class is \textit{airplane}. 
    $\lambda$ is fixed as 0.75.}  
    \label{tab:beta_d_1}
\end{table*}

\begin{table*}[htb!]
    \centering
    \begin{tabular}{r c c c c}
    \toprule
     $\beta_\text D \rightarrow$ &
     \multirow{2}{*}{0.01} & 
     \multirow{2}{*}{0.05} &
     \multirow{2}{*}{0.10} &
     \multirow{2}{*}{0.20} \\
     Poisoning rate $\downarrow$ & & & & \\
     \midrule
     10\% & \textit{acorn} & \textit{acorn} & \textit{acorn} & \textit{acorn} \\ 
     15\% & \textit{acorn} & \textit{acorn} & \textit{acorn} & \textit{acorn} \\ 
     20\% & \textit{acorn} & \textit{acorn} & \textit{acorn} & \textit{acorn} \\ 
    \bottomrule
    \end{tabular}
    \caption{Results of target classes detection for BadNets attack on ImageNet-30 for different values of hyperparameter $\beta_\text D$. The true target class is \textit{acorn}. $\lambda$ is fixed as 0.75.} 
    \label{tab:beta_d_2}
\end{table*}

\begin{table*}[htb!]
    \centering
    \begin{tabular}{r c c c c}
    \toprule
     $\beta_\text D \rightarrow$ &
     \multirow{2}{*}{0.65} & 
     \multirow{2}{*}{0.75} &
     \multirow{2}{*}{0.85} &
     \multirow{2}{*}{0.95} \\
     Poisoning rate $\downarrow$ & & & & \\
     \midrule
     0.65\% & \textit{airplane} & \textit{airplane} & \textit{airplane} & \textit{airplane} \\ 
     2.50\% & \textit{airplane} & \textit{airplane} & \textit{airplane} & \textit{airplane} \\ 
    \bottomrule
    \end{tabular}
    \caption{Results of target classes detection for LC attack on CIFAR-10 dataset for different 
    values of hyperparameter $\lambda$. The true target class is \textit{airplane}.
    $\beta_\text D$ is fixed as 0.05}  
    \label{tab:lambda_1}
\end{table*}

\begin{table*}[htb!]
    \centering
    \begin{tabular}{r c c c c}
    \toprule
     $\beta_\text D \rightarrow$ &
     \multirow{2}{*}{0.65} & 
     \multirow{2}{*}{0.75} &
     \multirow{2}{*}{0.85} &
     \multirow{2}{*}{0.95} \\
     Poisoning rate $\downarrow$ & & & & \\
     \midrule
     10\% & \textit{acorn} & \textit{acorn} & \textit{acorn, \textcolor{red}{mosque}} & \textit{acorn, \textcolor{red}{mosque}} \\ 
     15\% & \textit{acorn} & \textit{acorn} & \textit{acorn} & \textit{acorn} \\ 
     20\% & \textit{acorn} & \textit{acorn} & \textit{acorn} & \textit{acorn} \\ 
    \bottomrule
    \end{tabular}
    \caption{Results of target classes detection for BadNets attack on ImageNet-30 for different values of hyperparameter $\lambda$. The true target class is \textit{acorn}. $\beta_\text D$ is fixed as 0.05.} 
    \label{tab:lambda_2}
\end{table*}

\begin{table*}[htb!]
    \centering
    \begin{tabular}{r c c c c c}
    \toprule
     $\beta_\text D \rightarrow$ &
     \multirow{2}{*}{0.4} & 
     \multirow{2}{*}{0.5} &
     \multirow{2}{*}{0.6} &
     \multirow{2}{*}{0.7} &
     \multirow{2}{*}{0.8} \\
     Poisoning rate $\downarrow$ & & & & & \\
     \midrule
     1\% & \textit{airplane} & \textit{airplane} & \textit{airplane} & \textit{airplane} & \textit{airplane} \\ 
     5\% & \textit{airplane} & \textit{airplane} & \textit{airplane} & \textit{airplane} & \textit{airplane} \\ 
     10\% & \textit{airplane} & \textit{airplane} & \textit{airplane} & \textit{airplane} & \textit{airplane} \\ 
     20\% & \textit{airplane} & \textit{airplane} & \textit{airplane} & \textit{airplane} & \textit{airplane} \\ 
    \bottomrule
    \end{tabular}
    \caption{Results of target classes detection for BadNets attack on CIFAR-10 dataset for different 
    values of hyperparameter $\beta_\text{ND}$. The true target class is \textit{airplane}.}  
    \label{tab:beta_nd_1}
\end{table*}

\begin{table*}[htb!]
    \centering
    \begin{tabular}{r c c c c}
    \toprule
     $\beta_\text D \rightarrow$ &
     \multirow{2}{*}{0.65} & 
     \multirow{2}{*}{0.75} &
     \multirow{2}{*}{0.85} &
     \multirow{2}{*}{0.95} \\
     Poisoning rate $\downarrow$ & & & & \\
     \midrule
     1\% & \textit{acorn} & \textit{acorn} & \textit{acorn} & \textit{acorn} \\ 
     5\% & \textit{acorn} & \textit{acorn} & \textit{acorn} & \textit{acorn} \\ 
    \bottomrule
    \end{tabular}
    \caption{Results of target classes detection for BadNets attack on ImageNet-30 for different values of hyperparameter $\beta_\text{D}$.  The true target class is \textit{acorn}. $\lambda$ is fixed as 0.75.} 
    \label{tab:beta_nd_2}
\end{table*}

\section{Relabeling accuracies}
\label{app:relabeling_acc}
Table~\ref{tab:gen_accuracies} 
evaluates our generative classifier 
against the original labels, 
as they were prior to poisoning.
We observe the lowest accuracy 
in cases of Adap-Patch and Adap-Blend attacks.
We attribute this discrepancy 
to the tendency of these attacks
to enhance the similarity between 
the clean samples of the target class
and poisoned samples 
within the latent space. 

\begin{table}[htb!]
    \centering
    \begin{tabular}{lrrrrrr}
    \toprule
& BadNets & Blend & WaNet & ISSBA & Adap-P & Adap-B \\
\midrule
Relabeling accuracy & 92.6    & 88.3  & 90.3  & 92.6  & 42.1   & 36.5  \\
     \bottomrule
    \end{tabular}
    \caption{Relabeling accuracies [\%] of non-disruptive attacks on CIFAR-10. 
    As stated,
    the relabeling occurs only if the attack is classified as non-disruptive. } 
    \label{tab:gen_accuracies}
\end{table}

\end{document}